# Object Detection and Tracking with Autonomous UAV

A. Huzeyfe DEMIR[2]  Berke YAVAS[2]  Mehmet YAZICI[2]
ahmethuzeyfe.demir@ogr.iuc.edu.tr  berke.yavas@ogr.iuc.edu.tr  mehmet.yazici@ogr.iuc.edu.tr

Dogukan Aksu[1]  M. Ali AYDIN[2]
dogukan.aksu@huawei.com  aydinali@istanbul.edu.tr

[1] AI Enablement Department, Huawei Turkey R&D Center, Istanbul, Turkey
[2] Computer Engineering Department, Istanbul University – Cerrahpasa Istanbul, Turkey

*Abstract--* **In this paper, a combat Unmanned Air Vehicle (UAV) is modeled in the simulation environment. The rotary wing UAV is successfully performed various tasks such as locking on the targets, tracking, and sharing the relevant data with surrounding vehicles. Different software technologies such as API communication, ground control station configuration, autonomous movement algorithms, computer vision, and deep learning are employed.**

## I. INTRODUCTION

The first unmanned aerial vehicle was developed in 1916 by British engineer Archibald Montgomery Low. It was produced in limited numbers for use in World War I [1]. Nations compete globally to make unique and better advances in defense technologies. The fact that the Unmanned Aerial Vehicle (UAV) can be used in different areas, that it has advantageous physical and technical features, and that it has a structure that is open to development are some of the features that make it important.

UAV is a product that is becoming more and more popular in many areas such as daily life, agricultural spraying and agriculture, natural disasters, search and rescue works, reconnaissance studies, transportation, military technology. Since the invention of UAVs, there has been a continuous development and increase in usage areas in parallel with the development of software and hardware. It is foreseen that UAVs will have a permanent place in human life soon, with the steps taken to carry payloads, to perform image processing, to move autonomously, to fly for a long time, to increase their physical advantages with improvements, and to develop artificial intelligence.

This work includes different software modules such as Application Programming Interface (API) communication, ground control station configuration, autonomous motion algorithms, computer vision, video broadcasting, and machine learning.

Other parts of the paper categorized as follows: a literature review, used material and methods, experimental results of test cases on each module, and conclusion, respectively.

## II. LITERATURE REVIEW

Studies concerning mathematical modeling, mapping, improvement of autonomous movement, and algorithmic studies for performance improvement for UAVs are present in the literature.

Ekmen and Aydogdu mentioned in their work which simulation program experiments by Harmel et al. using the back-stepping control method to develop an UAV stability system, Bouabdallah et al. have tried multiple control methods on the UAV and successfully applied the proportional-integral-derivative (PID) control method on the Operating System 4 (OS4) experiment to achieve a stable flight, Kutluk Bilge Arıkan and his team to obtain a stable flying system that provides height control with linear-quadratic regulator (LQR) and orientation control with PID control algorithm [2].

Yilmaz, Aydin, and Cetinkaya, in their study on object detection using deep-learning, tested various deep-learning algorithms on their data set of 507 cup images and concluded that the Region-Based Convolutional Neural Network (R-CNN) algorithm produced the best results [3].

Dario Cazzato and his friends' study on computer vision with different object detection networks has shown that with the advent of deep learning, impressive results have emerged for real-world applications. Further, this study led us to conclude that a keyway to improve the performance of object detectors is to have enough datasets containing labeled images [4].

Mahmoud, Mohamed, and Al-Jaroodi's study has shown that the use of cloud technology through a server-client architecture and communication via web protocols in UAVs enable, through various applications and interfaces, easier and more flexible access to the vehicle, compared to other systems [5].

Liao and Juang similarly integrated the onboard computer, video streaming server, internet communications, Mongo database, and Kafka connector in their study for the detection of marine pollution [6].

Victor M. Becerra researched the detection of vehicles and other objects on the ground. Victor shows that all conditions must be taken seriously to lead to the success of the system. These conditions include the size of UAVs, wind conditions, battery, and many more [7].

Siyi Li and Dit-Yan Yeung, in their study on visual object tracking for unmanned aerial vehicles, showed that using an object tracker to track an object improves performance and improves tracking during sudden maneuvers [8].

The main contribution of our paper is to increase the effectiveness of UAVs in different areas of use such as defense industry, protection, search and rescue, criminal operations, by instantaneously tracking objects and sending data.

## III. MATERIAL AND METHODS

Autonomous movement, image processing and server-client modules are explained respectively in this section.

### A. Communication of Processes in the UAV

There are 4 processes running inside the UAV. These processes are Autonomous Node, Image Processing Node, Proxy, and Message Queuing Telemetry Transport (MQTT) Broker as shown in Figure 1. MQTT Broker is an intermediate component that broadcasts messages sent by 3 other applications. Other processes broadcast various messages to each other. These messages are distributed to other processes via MQTT Broker. MQTT Broker works with publish/subscribe model. The publishers and subscribers never contact each other directly.

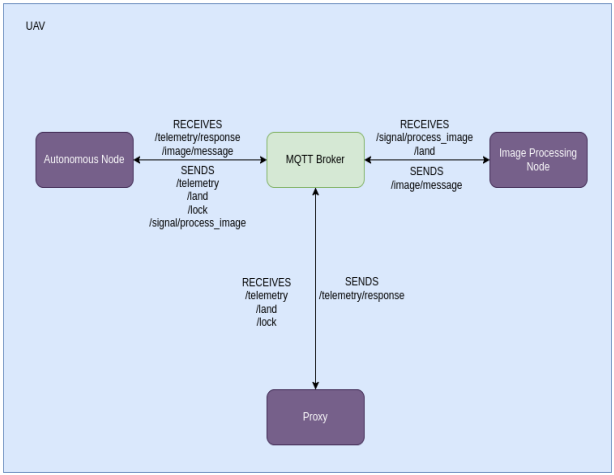

Fig. 1. Communication Diagram in UAV

When the UAV starts it, autonomous movement, it transmits a message to the Proxy via MQTT Broker with /telemetry topic. The proxy forwards this message to the Remote server and forwards the response to the Autonomous Node with the topic /telemetry/response. All applications are subscribed to the /land message that the Autonomous Node broadcasts. When this message arrives, the UAV has completed its autonomous movement and landed. This message transmits the signal to other applications to terminate their execution. Autonomous Node calculates the distance between itself and the target in the message it receives over /telemetry/response. If the distance is less than 10 meters, it broadcasts a message with the topic /signal/process_image. This message is read from the Image Processing Node and the image processing starts. After detecting the object during image processing, an X, Y message is published over the /image/message topic, indicating how the autonomous movement should continue. Kimon P. Valavanis explains the aspects of vision-based navigation and target tracking for UAVs in the book. He shows how to track a maneuvering ground target using a particle filter algorithm [9]. The Autonomous Node reads this message and adjusts the yaw and pitch values to get the target closer to the center of the camera. If the target stays inside the camera image for more than 10 seconds, a locking will occur. This time tracking takes place within the Autonomous Node. At the end of these 10 seconds, a message will be published via the /lock topic. This message is read over the proxy application and the locking information is transmitted to the remote server.

The diagram below shows the working principle of the Autonomous Node. The blue part represents the UAV take-off and the initiation of internal structures. After this part is completed, the UAV goes into SEARCH state. The SEARCH state is shown in the diagram in purple. In this part, the UAV exchanges data with the remote server and starts its movement towards the target. In case of data coming from the camera, the UAV switches to LOCK status. This diagram includes the orange parts on it. In the LOCK state, the UAV performs its autonomous movement with the data from the camera and starts a timer. If there is data from the camera for 10 seconds, it terminates the LOCK state and forwards the locking data to the remote server. It goes into the SEARCH state again and goes to the next target, if any, otherwise it goes into the landing state and broadcasts the land message.

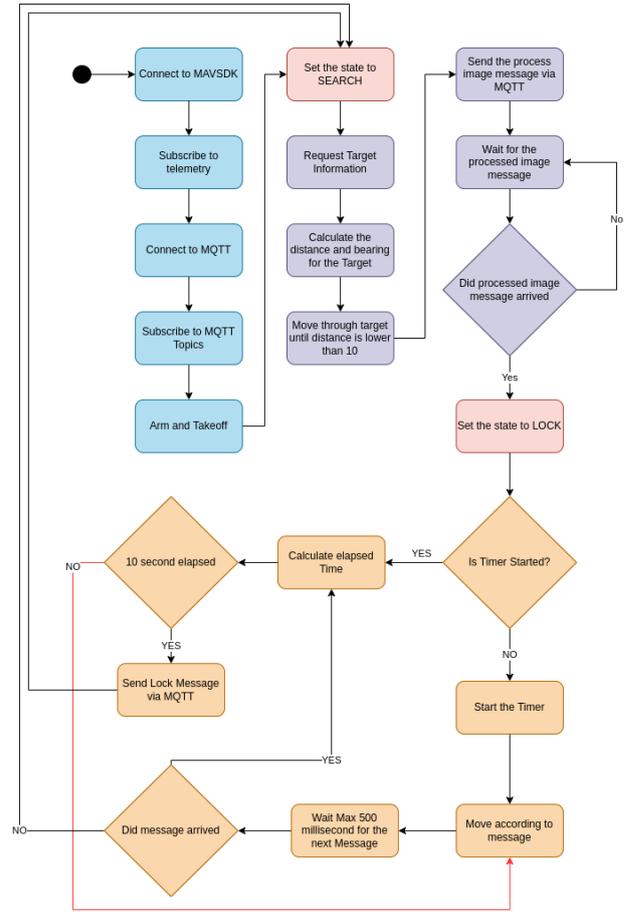

Fig. 2. Principle of Autonomous Node

### B. Image Processing Module

Python language was preferred as the software language in the image processing part of the project. Open-Source Computer Vision (OpenCV) was used as the image processing library. It provides parallel processing over Graphics Processing Unit (GPU). In this way, higher performance can be obtained when running difficult algorithms. Compute Unified Device Architecture (CUDA) technology is used to increase the parallel processing capability of the computer. You Only Look Once (YOLO) was used as an algorithm for object detection. YOLO is an algorithm for object detection using Convolutional Neural Networks (CNN). When the YOLO algorithm starts working, it can simultaneously detect objects in images or videos and their coordinates. Since the image processing in the project real-time, the number of frames processed per second should be high with minimum delay. YOLO was preferred in this project because it works better than similar algorithms and gives a higher FPS value. The model is trained with the Darknet framework. Darknet is an open-source neural network framework. YOLO is a heavy algorithm. Once we find the target object, we need to reduce the delay in order not to miss the target while following it. Therefore, it is planned to use an Object Tracker to improve performance. Because it is usually more difficult to detect an object from scratch and consumes more resources than tracking it. YOLO works until it detects the target object for the first time. Afterwards, the Channel and Spatial Reliability Tracking (CSRT) object tracker was used. This filter is used to search for the object in consecutive frames using the last known position of the object.

Fig. 3. Image Processing Schema

## C. Server-Client Module

During the duty period, the vehicle has communication with the remote server. Time, target location information, crash information is recorded on the remote server. Some of these data affect the movement, and some are the data obtained and recorded as a result of the movement. The vehicle has certain motion states. From the beginning of the movement, the vehicle is in the search state until the target vehicle is captured by the camera. In this state, the vehicle performs the selection of the target and obtaining the location information with the data obtained by the server calls. When the target is captured by the camera, the vehicle enters the locking state. In the locking state, the vehicle sends data packets containing information about the locking to the server and these data are recorded in the database.

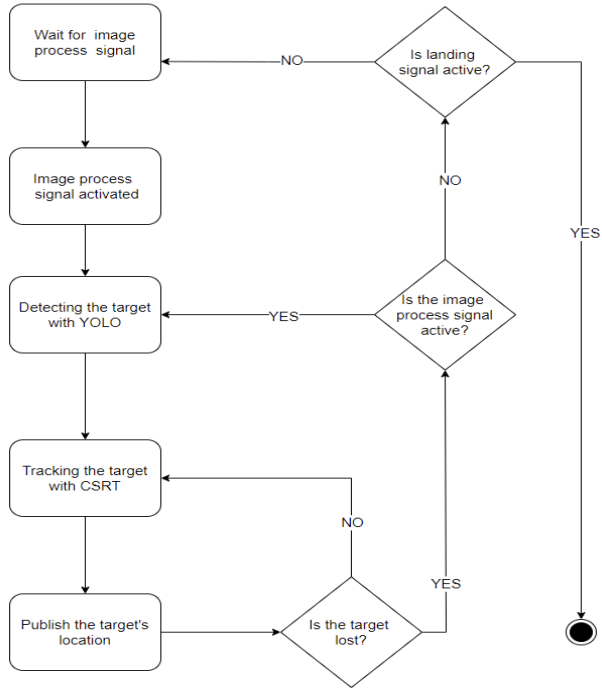

The server is developed with the Java Spring Boot framework. It is designed in accordance with Spring Model-View-Controller (MVC) architecture. This choice was made because it has a high rate of preference for creating Representational State Transfer (RESTful) API, has a quality design code base and a structure that provides sustainable development, and has a lot of features and packages. In-memory type H2 database has been preferred due to its easy implementation, easy portability, and low data load. The client was developed using the Go http package. Data transmission takes place via the HyperText Transfer Protocol (HTTP). Also in this module, there is a message transmission section developed with the Go MQTT package for the exchange of data with the remaining vehicle system. Go provides better performance than many languages. For this reason, Go was preferred because it is lighter and more efficient on the client side.

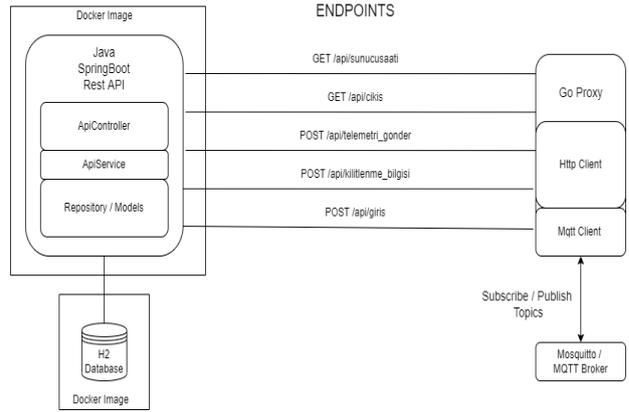

Fig. 4. HTTP Server/Client Architecture

## IV. EXPERIMENTAL RESULTS

Determining the conditions for the success of the project and the criteria for testing these conditions are important for the correct execution of the testing phase.

### A. System and Movement Tests

Pixhawk4 (PX4) autopilot was used for system test and Gazebo11 was used as simulation environment. In the photo below, the detection and tracking of the target autonomously during the simulation can be seen. In addition, the broadcast message is displayed on the terminal. The UAV is in the LOCK state in the photo below and continues its movement with the message displayed on the terminal.

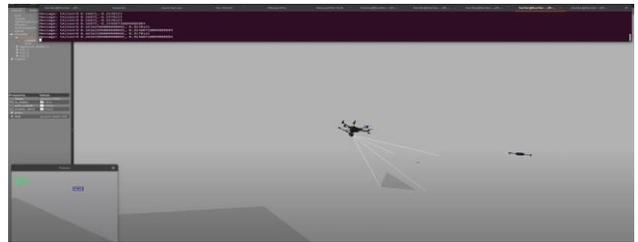

Fig. 5. System Tests on Gazebo

### B. Image Processing Module Tests

While testing the success of the model, it was tested for picture, video, and live video, respectively. The success of the training has been tested with parameters such as precision, recall, and F1-score.

### C. RESTful API Tests

The endpoints of the server were tested with the Postman application, the testing tool for the API. At this point, the accuracy of the requests and responses were checked. Data is transferred in JavaScript Object Notation (JSON) type. Afterwards, client tests were also combined with the server. In addition, in parallel with the project development process, test coverage increased.

### D. Results

In system and movement tests, it has been observed that the response of the UAV in scenarios where there are objects that do not move in both locking and search states during autonomous movement may need improvement in terms of performance and accuracy. There is no problem in detecting and tracking the variable or constant acceleration target UAV, and it has been observed that the algorithm does not work when the targets remain motionless in the air. The X, Y messages broadcast over the camera failed to follow the stationary targets, since these messages were constantly changing as the UAV was captured in motion, and these messages were used in the pitch and yaw part of the UAV autonomous movement.

The image processing module was trained with model 1359 UAV photos for tests. The target object was successfully detected in the picture, video, and live video tests of the model. The results we obtained as a result of model training are as follows.

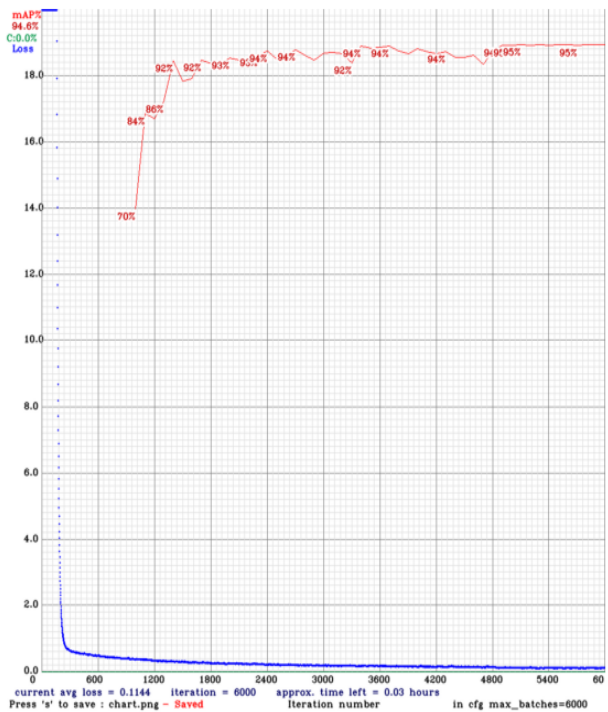

Fig. *6*. Success of Training

| Measure | Value | Definition |
|---|---|---|
| Iterations | 6000 | Total number of iterations. |
| Avg Loss | 0.1144 | The average of the errors in the model. |
| TP | 150 | True Positive |
| FP | 9 | False Positive |
| FN | 14 | False Negative |
| Precision | 0.94 | TP / (TP + FP) |
| Recall | 0.91 | TP / (TP + FN) |
| F1-score | 0.93 | 2TP / (2TP + FP + FN) |

Table 1. Training Results

In the server endpoint tests, the connection controls in the communication and the integrity of the incoming data were checked. At this stage, it was observed that the requests made were successful and the incoming responses were as requested. In addition, it has been observed that the latency of requests is 95 ms for an average of 500 bytes of data on the local machine.

## V. CONCLUSION

The system is designed and written to process the information of the target that the UAV receives from the remote server, proceed autonomously to the coordinates of the target, detect the relevant target via the camera and report it back to the server. The developments in the project can be adapted to all kinds of scenarios where the performer of all kinds of assets and actions in areas where human cannot reach in a short time needs to be detected autonomously and this information needs to be reported to a remote server.

The desired success was achieved in the tests performed with simulation. However, in the construction of a physical UAV, many different issues can be added to the business planning. The success rate may be adversely affected due to technical problems such as materials and inventory to be used, connection errors, unfavorable environment, insufficient energy supply. Since the simulation world provides ideal conditions and there is no noise in the image formed in the camera module, it is thought that higher success has been achieved compared to the tests of the software in a vehicle in the physical environment. For this reason, it may be necessary to produce both hardware and software solutions in case such problems occur during the operation of the software on a UAV. For example, noise in the image can be reduced by using a high-resolution camera to achieve the success achieved in the simulation.

It has been observed that the algorithm fails if the targets are suspended motionless in the air. This is because the active velocity of the target detected only from the camera is unknown. If a sensor that detects the speed of the target is added to the system, the system can adjust the speed with the data coming from this sensor, and thus it can be performed on objects that are motionless in the air.

Since the remote server and the UAV were on the same network during the simulation, there was no interruption in communication and success was achieved with minimum delay. In real life, disconnections or delays are expected during the communication between the server and the UAV.

In the image processing part, the FPS value was tried to be increased by processing with GPU. The model can be trained with more UAV images to improve image detection performance. Dawei Du and his friends' study on object detection and tracking for the unmanned aerial vehicle has shown that small objects are more difficult to detect, so the dataset needs to be improved to get better results [10]. Another option would be to use a GPU with higher processing power and increase the resolution used for image processing. Various visual disturbances (rain, insufficient light, light reflections, etc.), noise in the image, shaking, etc. that may occur in the physical realization of the project. These factors will reduce image processing performance. Some noise canceling filters can be used to solve this. When the model training results are examined, the tests made with pictures, videos and live videos were found to be successful. Average loss value was targeted to be below 30 % and 11.44 % was obtained. Precision and recall values above 90% were found to be sufficient.

ACKNOWLEDGEMENTS

This work is also a part of the B.Sc. thesis titled Rotary Wing Fighting Unmanned Aerial Vehicle (UAV) Software Systems Development and Modeling in Simulation Environment Istanbul University, Faculty of Engineering.

REFERENCES

[1] V. Prisacariu, *The History and The Evolution of UAVs From the Beginning Till The 70s,* Journal of Defense Resources Management, vol. 8, no. 1, pp. 181-189, 2017

[2] M. I. Ekmen, O. Aydogdu, *Image Processing Based Autonomous Landing for Unmanned Aerial Vehicles*, European Journal of Science and Technology, Special Issue, pp. 297-303, Sept. 2020

[3] M. Koc, *Derin Öğrenme Kullanarak İHA ile Hareketli Bir Hedefin Otonom Olarak Yakalanması*, Department of Computer Engineering, Sabahattin Zaim University, Istanbul, 2020

[4] D. Cazzato, C. Cimarelli, J. L. Sanchez-Lopez, H. Voos, M. Leo, *A Survey of Computer Vision Methods for 2D Object Detection from Unmanned Aerial Vehicles,* Journal of Imaging, vol. 6, no. 8, pp. 78, Aug. 2020

[5] S. Mahmoud, N. Mohamed, J. Al-Jaroodi, *Integrating UAVs into the Cloud Using the Concept of the Web of Things,* Hindawi Publishing Corporation Journal of Robotics, vol. 2015, no. 631420, Oct. 2015

[6] Y. Liao, J. Juang, *Real-Time UAV Trash Monitoring System*, Applied Sciences, vol. 12, pp. 1838, 2022

[7] K. P. Valavanis, *Advances in Unmanned Aerial Vehicles: State of the Art and the Road to Autonomy*, Springer, Florida, USA, 2007

[8] S. Li, D. Yeung, *Visual Object Tracking for Unmanned Aerial Vehicles: A Benchmark and New Motion Models*, AAAI, vol. 31, no.1, pp. 4140-4146, 2017

[9] V. M. Becerra, *Autonomous Control of Unmanned Aerial Vehicles*, Electronics, vol. 8, no. 4, pp. 452, 2019

[10] D. Du, Y. Qi, H. Yu, Y. Yang, K. Duan, G. Li, W. Zhang, Q. Huang, Q. Tian, *The Unmanned Aerial Vehicle Benchmark: Object Detection and Tracking*, ECCV, Computer Vision, pp. 375–391, 2018